\documentclass[portugues]{sobraep}

\usepackage{algpseudocode}
\usepackage{algorithm}
\usepackage[bottom]{footmisc}

\DeclareMathOperator*{\argmax}{\arg\!\max}
\DeclareMathOperator*{\argmin}{\arg\!\min}
\DeclareMathOperator{\E}{\mathbb{E}}
\DeclareMathOperator{\D}{\mathcal{D}}
\DeclareMathOperator{\KL}{D_{KL}}
\DeclareMathOperator{\bpi}{\pi_\theta}
\DeclareMathOperator{\lpi}{\pi_\vartheta}

\titulo{Latent-Variable Advantage-Weighted Policy Optimization for Offline RL}

\author{Xi Chen$^{1}$, Ali Ghadirzadeh$^{2}$, Tianhe Yu$^{2}$, Yuan Gao$^{3}$, Jianhao Wang$^{1}$,\\Wenzhe Li$^{1}$, Bin Liang$^{1}$, Chelsea Finn$^{2}$, Chongjie Zhang$^{1}$ \\ $ $ \\
	\normalsize $^{1}$ Tsinghua University, China, $^{2}$ Stanford University, USA, $^{3}$ Shenzhen Institute of Artificial Intelligence and Robotics for Society, CUHK Shenzhen, China   
}

\begin{document}

\maketitle

\begin{abstract}
Offline reinforcement learning methods hold the promise of learning policies from pre-collected datasets without the need to query the environment for new transitions. This setting is particularly well-suited for continuous control robotic applications for which online data collection based on trial-and-error is costly and potentially unsafe. In practice, offline datasets are often \emph{heterogeneous}, i.e., collected in a variety of scenarios, such as data from several human demonstrators or from policies that act with different purposes. Unfortunately, such datasets can exacerbate the distribution shift between the behavior policy underlying the data and the optimal policy to be learned, leading to poor performance. To address this challenge, we propose to leverage latent-variable policies that can represent a broader class of policy distributions, leading to better adherence to the training data distribution while maximizing reward via a policy over the latent variable. As we empirically show on a range of simulated locomotion, navigation, and manipulation tasks, our method referred to as latent-variable advantage-weighted policy optimization (LAPO), improves the average performance of the next best-performing offline reinforcement learning methods by 49\% on heterogeneous datasets, and by 8\% on datasets with narrow and biased distributions. 
\end{abstract}

\begin{figure*}[ht]
\begin{center}
\centerline{\includegraphics[width=2\columnwidth, bb=0 0 1350 450]{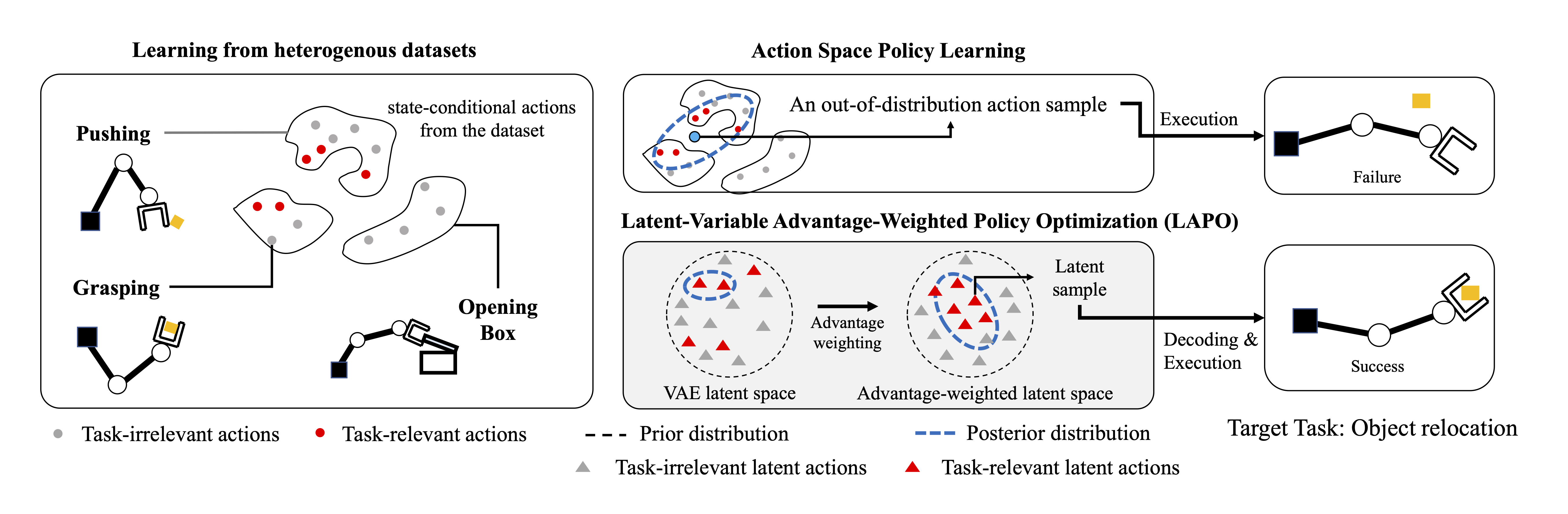}}
\caption{An example scenario of learning from data with heterogeneous action distributions. (left) The dataset includes data from three different tasks, pushing, grasping, and opening boxes. (right) The learned task is to relocate an object to a goal position. (middle up) Learning policies directly in the action space can result in sampling from out-of-distribution actions which subsequently can fail the learning task. (middle bottom) LAPO constructs a state-conditional latent space that maps actions that lead to high rewards to the same neighborhood, and learns a latent policy that can capture task-relevant actions without including out-of-distribution samples. }
\vspace{-0.5 cm}
\label{fig:lapo_overview}
\end{center}
\end{figure*}

\section{Introduction}
\label{sec:intro}
Offline reinforcement learning (RL), also known as batch RL \cite{lange2012batch}, addresses the problem of learning an effective policy from a static fixed-sized dataset
without interacting with the environment to collect new data. This formulation is especially important for robotics, as it avoids costly and unsafe trial-and-error and provides an alternative way of leveraging a pre-collected dataset. However, in practical settings, such offline datasets are often heterogeneous and are collected using different policies, leading to a data distribution with multiple modes.
These data-collection policies may aim to accomplish tasks that are not necessarily aligned with the target task 
or may accomplish the same task but provide conflicting solutions. 
In contrast to the prior works that have focused on the distributional shift problem 
\cite{kumar2019stabilizing, fujimoto2019off, kumar2020conservative}, here, we address the problem of learning from heterogeneous data.

One main challenge of learning from heterogeneous datasets is the existence of conflicting actions in the dataset that lead to high rewards. For example, if the target task involves solving two primitive tasks in a sequence and the action distributions of the two tasks are orthogonal due to the heterogeneity of the dataset (as shown in Figure~\ref{fig:lapo_overview}), the learned policy might generate out-of-distribution actions that do not lie in either action space of the two primitives, exacerbating the distributional shift problem.
This is particularly challenging for implicit policy constraint methods that formulate a supervised learning objective function and the forward Kullback–Leibler (KL)-divergence between the parametric policy being learned and the closed-form optimal policy found through advantage-weighted behavior cloning  \cite{nair2020awac, peng2019advantage, siegel2020keep}. 
As shown in Figure~\ref{fig:lapo_overview}, state-conditional distributions learned directly over the action space using the forward KL-divergence objective may assign high probability to out-of-distribution actions and, as we empirically show, can result in sub-optimal policies with heterogeneous datasets.
A similar problem has been reported when explicitly constraining the policy on multi-modal distributions using Maximum Mean Discrepancy (MMD) distances \cite{zhou2020plas}.

Figure~\ref{fig:lapo_overview} also illustrates a potential solution to the problem of learning from heterogeneous datasets containing conflicting actions. 
The intuition is to construct a latent space in which, conditioned on a state, actions that lead to high rewards are mapped to the same neighborhood. Then, we can learn simple state-conditional distributions such as Gaussian distributions over the latent space which captures task-relevant and high-reward actions without including out-of-distribution samples.
Based on this intuition, we propose to learn a latent-space policy by alternating between  training the policy and maximizing the advantage-weighted log-likelihood of data.
This biases the RL policy to choose actions that are both supported by the training data and effective for the target task. 

The main contributions of this work 
is the introduction of a new method, which we refer to it as latent-variable advantage-weighted policy optimization (LAPO), that can efficiently solve heterogeneous offline RL tasks. 
LAPO  learns an advantage function and a state-conditional latent space in which high-advantage action samples are generated by sampling from a prior distribution over the latent space. Furthermore, following a prior work, \cite{zhou2020plas}, we also train a latent policy that obtains state-conditioned latent values which result in higher reward outcomes compared to latent samples directly drawn from a prior distribution. 
We compare LAPO to vanilla behavior cloning, BCQ \cite{fujimoto2019off}, PLAS \cite{zhou2020plas}, AWAC \cite{nair2020awac}, IQL \cite{kostrikov2021offline_iql}, and CQL \cite{kumar2020conservative} on a variety of simulated locomotion, navigation, and manipulation tasks, provided heterogeneous offline datasets and also biased datasets with narrow data distribution in standard offline RL benchmarks. 
LAPO is the only method that yields good performance across all of these tasks and on average improves by 49\%  over the next best method for heterogeneous datasets, and by 8\% on other offline RL tasks with narrow data distributions. 

\section{Preliminaries}
\label{sec:preliminaries}
\subsection{Offline Reinforcement Learning}
The goal of reinforcement learning is to obtain a policy that maximizes a notion of accumulated reward for a task as a Markov decision process (MDP). 
In offline RL settings, we assume that we are given a dataset $\D$ containing trajectories of state-action pairs $ \{(s_0, a_0), ..., (s_T, a_T) \} \in \D$,
where, $s_t \in \mathcal{S}$ and $a_t \in \mathcal{A}$ denote the state and action at the time step $t$. We assume that all states in the dataset are sampled from a fixed transition probability distribution $p(s_{t+1} | s_t, a_t)$ while actions can come from a mixture of policies which is referred to as the behavior policy. 
The policies may try to accomplish different tasks and hence generate trajectories unrelated to the target task.

The RL task is specified by an MDP with the same state and action spaces and transition probability distribution (which is unknown to the learning agent). We assume that we are given a task reward function $r(s, a)$ that assigns a bounded reward value to every state-action pair, and a discount factor $\gamma \in [0,1]$ which is used to formalize the expected discounted reward objective function, 
\begin{equation}
    \label{eq:rl_objective}
    J(\theta) = \E_{ s_t \sim p_{\pi}, a_t \sim \pi_\theta} [\sum_{i=t}^T \gamma ^ {i-t} r(s_i, a_i)],
\end{equation}
where, $\pi_\theta(a_t | s_t)$ denotes the parametric policy, with parameters $\theta$, being learned. 
The action-value function, or Q-function, of the policy is defined as $Q^\pi(s,a) = \E_\pi[\sum_{i=0}^T \gamma^i r(s_{t+i}, a_{t+1}) | s_t = s, a_t = a]$, and for a parameterized Q-function (with parameters $\phi$) implemented by a neural network, can be trained using the following learning rule:
\begin{equation}
\label{eq:policy_evaluation}
\begin{split}
    \argmin_{Q_\phi} \E_{ \D} [&(r(s_t, a_t) + \gamma V(s_{t+1}) \\ 
                            &- Q_{\phi}(s_t, a_t))^2],
\end{split}
\end{equation}
where, $V$ denotes the value function $V(s) = \E_{\pi_\theta(a|s)} [Q_{\phi'}(s, a)]$ and can be approximated by sampling actions from the policy distribution and averaging the corresponding action-value. $\phi'$ denotes the parameters of the Q-function at the previous iteration. 
The advantage of a pair of state and action is then defined as $A(s, a) = Q_\phi(s, a) - V(s)$. 

\subsection{Implicit Policy Constraints with KL-divergence}
A central challenge for offline RL methods is to limit the state-conditional action distribution of the learned policy to the empirical conditional action distribution of the dataset. 
It is very likely for off-policy algorithms to have overly optimistic values, as they tend to query the Q-function for action inputs outside the training distribution. 
Policy constraints methods \cite{kumar2019stabilizing, wu2019behavior} deal with this problem by keeping the learned policy close to the behavior policy  using the following constrained policy optimization formulation:  
\begin{equation}
    \begin{split}
        \argmax_{\pi} & \E_{s \sim \D} \E_{\pi(a|s)}[A(s,a)], \\
        & \text{s.t.} \, \E_{s \sim \D}[D(\pi_\theta(a|s), \pi_\beta(a|s))] < \epsilon
    \end{split}
    \label{eq:constrained_policy_opt}
\end{equation}
where, $\pi_\beta(a|s)$ denotes the unknown empirical conditional action distribution of the dataset, and $D$ denotes a distance measure such as KL-divergence \cite{nair2020awac, siegel2020keep} or  maximum mean discrepancy (MMD)  \cite{kumar2019stabilizing}, and $\epsilon$ is a threshold parameter. 
In case, we use the KL-divergence as the divergence in Equation \ref{eq:constrained_policy_opt}, the optimal $\pi^*$ can be expressed as 
\begin{equation}
    \pi^*(a|s) \propto \pi_\beta(a|s)\exp(A(s,a)/\lambda), 
    \label{eq:optimal_policy}
\end{equation}
where, $\lambda$ is a temperature parameter that depends on the $\epsilon$. 

Prior work \cite{nair2020awac, siegel2020keep, peng2019advantage} suggested to incrementally solve Equation \ref{eq:optimal_policy} by representing the optimal policy $\pi^*(a|s)$ as a non-parametric policy, and then project it onto the parametric policy $\pi_\theta(a|s)$ via supervised regression:
\begin{equation}
\begin{split}
\argmin_\theta \E_{\D}[\KL(\pi^*(a|s) || \pi_\theta(a|s))] = \\
\argmax_\theta  \E_{\D}[\,\E_{\pi^*(a|s)} [\log \pi_\theta(a|s)]\,],
\end{split}
\label{eq:forward_kl}    
\end{equation}
where, the expectation is estimated by sampling from the dataset. 
In practice, the non-parametric policy can be implemented by weighting samples from the dataset $\D$ using importance weights $\omega = \exp(A(s,a)/\lambda)$, and the projection via KL-divergence can be done via weighted supervised learning using these weights \cite{levine2020offline}. 

\section{Latent-Variable Advantage-Weighted Policy Optimization}
\label{sec:method}
In this work, we aim to address the problem of learning a policy from a heterogeneous offline dataset whose distribution has multiple modes, e.g., because of data collected by several human demonstrators or by policies with different purposes. We first start by motivating why this setting is challenging for implicit policy constraint methods. We then introduce our method that addresses this problem. 
\begin{figure}[h]
\vskip 0.2in
\begin{center}
\centerline{\includegraphics[width=1.0\columnwidth]{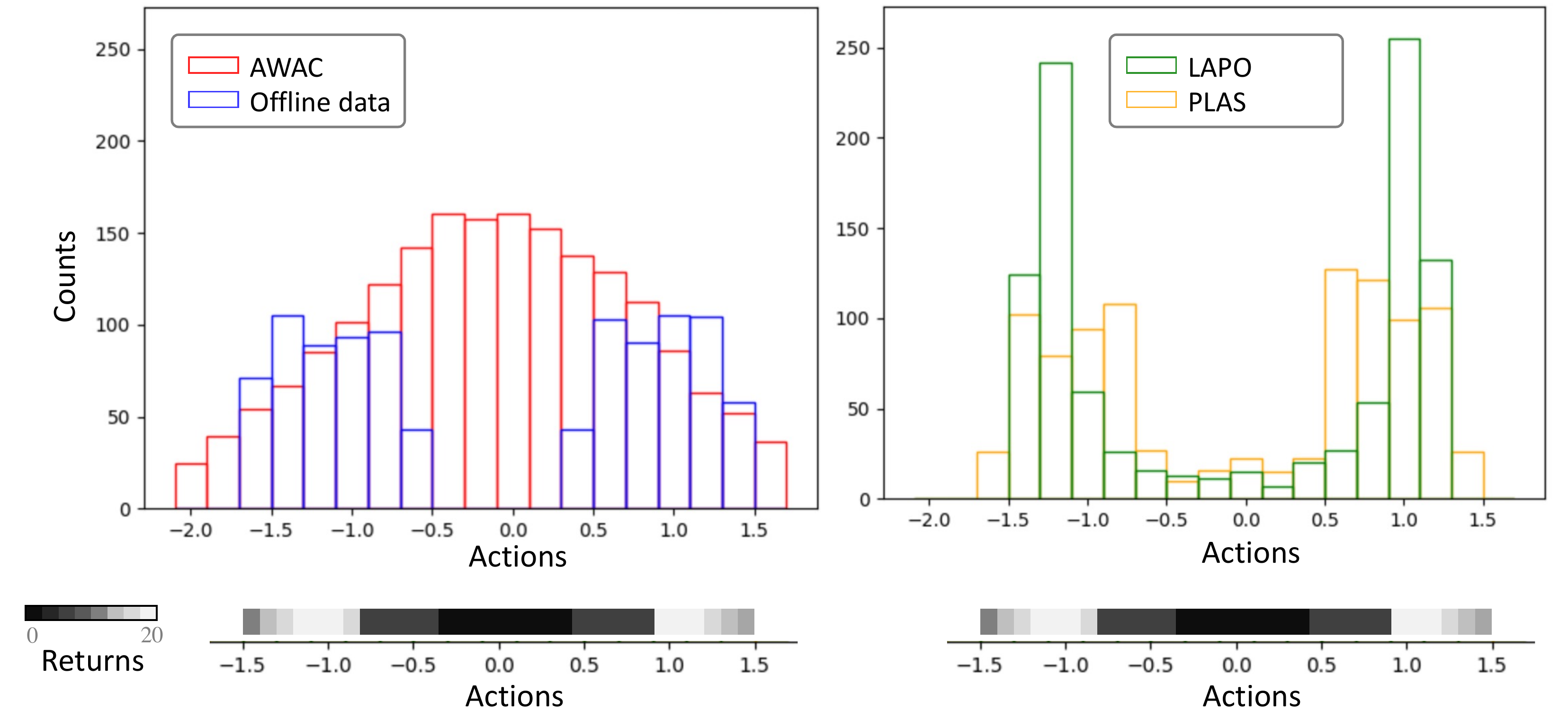}}
\caption{The histogram of actions taken at the initial state in the offline dataset (in blue), the histogram of actions sampled from the state-conditional action distribution learned by the AWAC method provided this offline dataset (in red), the histogram of actions learned by the PLAS generative model (in yellow), and the histogram of actions learned by the LAPO generative model (in green). 
The majority of the actions sampled from AWAC are out-of-distribution and fail the agent to complete the task, while the majority of actions generated by the latent model are  in-distribution. Besides, compared to PLAS, LAPO generates a significantly larger number of high-reward actions.}
\label{fig:forward_kl}
\end{center}
\vskip -0.2in
\end{figure}

The main challenge in learning from heterogeneous datasets results from optimizing an objective function based on the forward KL divergence to represent data distributions with multiple modes. 
As an example, consider a simple navigation task in which an agent navigates to a goal position while avoiding obstacles. In this example, the demonstrated expert actions have two prominent modes at the initial state corresponding to the actions moving to either the right or left side of the obstacle. 
The histogram of actions at the initial state in Figure~\ref{fig:forward_kl} shows the two modes.  
For Gaussian policy distributions, the optimal policy  $\pi_\theta(a|s)$ for the objective function given by Equation~\ref{eq:forward_kl} should assign high probability wherever the optimal policy $\pi^*(a|s)$ has high probability. As shown in  Figure ~\ref{fig:forward_kl}, the learned probability distribution, depicted in red, covers the two modes of the optimal policy but inevitably assigns high probability on out-of-distribution actions which in our example results in a collision with the obstacle.  
A similar problem has also been reported by the prior work \cite{zhou2020plas} for objectives constructed based on the MMD distance for bi-modal action distributions. 
Therefore, policy constraint methods based on KL divergence or MMD distance may be ineffective for offline RL problems with heterogeneous datasets.  

To address the challenges described above, we present a novel method, named {\em latent-space advantage-weighted policy training} (LAPO). At a high-level, LAPO has two key components. First, LAPO learns a latent-variable \emph{action policy} $\bpi(a|s, z)$, with $z$ being the latent variable, by regressing the actions in the dataset, weighted by the estimated advantage of each action. This first component extends advantage-weighted regression (AWR)~\cite{peng2019advantage} with a more expressive policy distribution class. In particular, by using a latent variable model, this policy class is substantially more expressive than a Gaussian action space.
Also, as illustrated in Figure~\ref{fig:forward_kl}, weighting by advantages allows it to effectively model the high-reward regions of data compared to regular latent-variable policy training \cite{zhou2020plas}, even when given data with multiple modes. 
Second, to improve the performance further, LAPO also learns a \emph{latent policy} $\lpi(z|s)$, which is trained to directly maximize the expected return using the latent action. As the action policy captures behaviors that are represented in the data, this latent policy will naturally avoid out-of-distribution actions and its learning will in turn improve the original action policy. Next, we will describe each of these components in detail.

\textbf{Updating the action policy:}
LAPO incrementally learns a latent-variable action policy $\bpi(a | s , z)$ by alternating between two steps: (1) estimating the advantage of each action, and (2) regressing the actions in the dataset, weighted by the estimated advantage of each action. 
The second step is similar to the advantage-weighted regression (AWR) for approximating the optimal policy $\pi^*$ in Equation~\ref{eq:optimal_policy}.  

In practice, the action policy is learned by maximizing the log-likelihood of the action data weighted by the exponential advantages $\omega = \exp(A(s,a)/\lambda)$. Following \cite{kingma2013auto}, we derive a weighted variational lower bound that maximizes the weighted log-likelihood of data: 
\begin{equation}
\label{eq:vae_objective}
\begin{split}
     \max_{\bpi, q_\psi} \E_{s,a \sim \D} [\omega 
    \E_{q_\psi(z|s, a)} &[\,\log (\bpi(a|s, z))\, - \\
    &\beta\KL(q_\psi(z|s, a)\, ||\, p(z))\,], 
\end{split}
\end{equation}
where $q_\psi(z|s,a)$ is an amortized variational distribution, which is an auxiliary model to approximate the conditional posterior distribution over the latent variable. 
$p(z)$ is the prior distribution over the latent variable and typically modelled by a standard normal distribution.
Finally, $\beta$ is a constant value that is used as a hyper-parameter to balance the two loss terms \cite{higgins2016beta}.   

\textbf{Policy evaluation:} 
As explained earlier, the estimated advantage of each state-action pair is used to update the action policy. These advantage estimations are incrementally found through a policy evaluation process in which the overall policy, formed by combining the action policy and the latent policy, is being evaluated. 
The state-conditional action distribution of the overall policy $\pi_{\theta, \vartheta}(a | s)$, is found by first sampling a latent value from the latent policy $z \sim \lpi(z | s)$, and then feeding the latent value to the action policy to sample the action $a \sim \bpi (a|s,z)$. 
In the policy evaluation step, we update the task action-value function by minimizing the squared temporal difference error using Equation \ref{eq:policy_evaluation}. 
To calculate the expected value of a state, we sample actions from the overall policy, i.e., 
\begin{equation}
    \label{eq:value_function}
    V(s) = \E_{ \substack{ a \sim \pi_\theta(. | s, z) \\ z \sim \pi_\vartheta(. | s) }} [Q_{\phi'}(s, a)].
\end{equation}
Since the latent values $z$ will be used as the input to the action policy, an unbounded $z$ can lead to generating out-of-distribution actions. Therefore, to better ensure sampling in-distribution actions, $\lpi(z|s)$ can be modeled as a truncated Gaussian \cite{brock2018large}. 
For deterministic versions of the latent policy, the policy output can be limited to $[-z_{\max}, z_{\max}]$ using a $\tanh$ function.  



\textbf{Policy improvement:}
LAPO optimizes the latent policy $\lpi(z | s)$ in every iteration to directly maximize the return. 
It is trained based on standard RL approaches such as DDPG \cite{lillicrap2015continuous} or TD3 \cite{fujimoto2018addressing} to maximize the RL objective in Equation \ref{eq:rl_objective}, by learning latent actions which result in high returns after being converted to the original action space using the action policy.  

The latent policy is updated by maximizing the Q-function over states sampled from the dataset, latent variables from the latent policy, and actions from the action policy:
\begin{equation}
    \label{eq:latent_policy_update}
    \argmax_{\pi_\vartheta} \E_{ \substack{  a \sim \pi_\theta(.|s,z) \\ z \sim \pi_\vartheta(.|s) \\ s \sim \D }} [Q(a,s)].
\end{equation}
Please note that, at this phase, the parameters of the action policy are fixed and are not affected by the update rule. 
For deterministic policies, $\bpi(s_t, z_t)$ and $\lpi(s_t)$, we can use the deterministic policy gradient algorithm to obtain the gradient of the objective function $J(\lpi)$ w.r.t. the parameters of the latent policy
\begin{equation}
\label{eq:policy_gradient}
\begin{split}
    \nabla_\vartheta J(\vartheta) = \E_{s \sim \D}[&\nabla_a Q(a, s)|_{a = \pi_\theta(s,z)} \\
    &\nabla_z\pi_\theta(s, z)|_{z=\pi_\vartheta(s)}\nabla_\vartheta\pi_\vartheta(s)]
\end{split}
\end{equation}

A summary of our method is presented in Algorithm \ref{alg:method}. 
We randomly initialize the parameters of the action-value function, the action policy, the amortized variational distribution, and the latent policy.
In every training iteration, we first update the action policy and the variational distribution $q$ using Equation \ref{eq:vae_objective} given the latest values of the advantage weights.  
Then, we update the action-value function using Equations \ref{eq:policy_evaluation} and \ref{eq:value_function} provided the updated action and latent policies. 
Then, for every state-action pair in the dataset, we compute the advantage using the updated action-value function, the behavior policy, and the latent policy, and then estimate the weights as the exponential of the scaled advantages, with the scale factor $\lambda$. 
Finally, we update the latent policy using Equation \ref{eq:latent_policy_update}, or, in case of deterministic policies, Equation \ref{eq:policy_gradient}. 

\begin{algorithm}[tb]
\caption{Latent-Variable Advantage-Weighted Policy Optimization}
\label{alg:method}
\begin{algorithmic}
\State Input: $\D = \{(s_t, a_t, s_{t+1})_i\}, i = 1, \dots, N$. 
\State Initialize: $\phi, \theta, \psi, \vartheta$ randomly, and $\omega = 1.0$. 
\For {$M$ iterations}
\State Update the behavior policy $\bpi$ using Equation \ref{eq:vae_objective}.
\State Update the Q-function $Q_\phi$ using Equations \ref{eq:policy_evaluation} and \ref{eq:value_function}.
\State Estimate the advantage weights for every state-action  
\State using $\omega = \exp(\frac{Q_\phi(s, a) - V(s)}{\lambda})$ 
\State Update the latent policy $\lpi$ using Equation \ref{eq:latent_policy_update}
\EndFor 
\end{algorithmic}
\end{algorithm}

\section{Related Works}
\textbf{Offline Reinforcement Learning:} Offline RL methods generally address the problem of distribution shift between the behavior policy and the policy being learned~\cite{fujimoto2019off,kumar2019stabilizing}, which can cause issues due to out-of-distribution actions sampled from the learned policy and passed into the learned critic. To address this issue, prior methods constrain the learned policy to stay close to the behavior policy via explicit policy regularization~\cite{liu2020provably,jaques2019way,wu2019behavior, kumar2019stabilizing, kostrikov2021offline,ghasemipour2021emaq}, via implicit policy constraints \cite{siegel2020keep, wang2018exponentially, peters2010relative, peng2019advantage, nair2020awac, kostrikov2021offline, kostrikov2021offline_iql}, by regularizing based on importance sampling \cite{nachum2019dualdice, liu2019off}, by learning of conservative value functions~\cite{kumar2020conservative,sinha2021s4rl}, by leveraging auxiliary behavioral cloning losses~\cite{fujimoto2021minimalist, nair2020accelerating}, and through model-based training with conservative penalties~\cite{yu2020mopo,kidambi2020morel,argenson2020model,swazinna2021overcoming,matsushima2020deployment,lee2020representation,yu2021combo}. Compared to these prior works, we opt to develop a method that uses an implicit policy constraint. However, prior implicit policy constraint methods have been largely limited to Gaussian policies, leading to poor performance on heterogeneous datasets, which can potentially exacerbate the distribution shift problem. Moreover, we find that AWAC~\cite{nair2020awac}, a representative approach in this category, struggles on heterogeneous data even when combined with policies parameterized by Gaussian mixture models (GMM). We overcome this challenge by introducing a new method that leverages latent variable models. As we will find in Section~\ref{sec:expts}, our method also outperforms state-of-the-art prior methods in other categories, particularly when learning from heterogeneous datasets.

Prior works~\cite{agarwal2021persim,kalashnikov2021mt,yu2021conservative} also studied the problem of learning from heterogeneous dataset in the offline RL with heterogeneous dynamics setting~\cite{agarwal2021persim} and multi-task offline RL with data sharing scenario~\cite{kalashnikov2021mt,yu2021conservative} respectively. Both settings require knowledge of the agent/task identifier.  Here, we study a more general problem setting including datasets with conflicting actions without assuming access to the ground-truth identifier of the source of the heterogeneity. 

\textbf{RL with Generative Models:}
Generative models have been used by prior works for improving training performance \cite{singh2020parrot, ghadirzadeh2017deep, ghadirzadeh2020data}, enabling transfer learning \cite{arndt2020meta, ghadirzadeh2021bayesian}, implementing hierarchical RL \cite{ajay2020opal, luo2020carl, lynch2020learning, peng2019mcp}, avoiding distributional shift in offline RL settings \cite{fujimoto2019off, zhou2020plas}, and learning dynamics models~\cite{lee2019stochastic,hafner2019learning,rafailov2021offline,hafner2019dream}. 
Our proposed method resembles PLAS \cite{zhou2020plas}, which learns a generative model and latent policy; but, unlike this prior work, our method trains an advantage-weighted generative model by alternating between learning the generative model, the advantage function, and the latent policy. This allows LAPO to capture different high-reward solutions using a simple Gaussian distribution in the latent space. As we will find in Section~\ref{sec:expts}, this distinction is crucial, as LAPO significantly outperforms PLAS on heterogeneous datasets.

\section{Experiments}
\label{sec:expts}
Our experiments aim to answer the following questions: 
(1) How does LAPO compare to other offline RL methods on a set of standard offline RL tasks, including learning from heterogeneous and homogeneous offline datasets? 
(2) How does LAPO compare to prior methods implemented with GMM policies when learning behaviors offline from heterogeneous datasets? 
and (3) In which setting does LAPO benefit from the latent policy training, versus only using the action policy? Can LAPO learn without constraining the latent values? 

We study the first question by evaluating LAPO and comparing it to several prior methods on offline RL benchmarks with heterogeneous datasets.
We consider a range of simulated robotic tasks, including navigation, locomotion, and manipulation, each with a corresponding static dataset with a heterogeneous data distribution.
We also consider datasets with narrow and biased data distributions, containing near-optimal or random trajectories. 
To answer the second question, we include a comparison of AWAC~\cite{nair2020awac} with a GMM policy, since AWAC trains using a similar advantage-weighted objective as the actor policy in LAPO. This comparison is only performed on heterogeneous datasets, since this is where we expect to see the most improvement from a multi-modal policy.
Finally, the questions in (3), we conduct two ablation studies in which we train LAPO without learning the latent policy, i.e. evaluating only LAPO's action policy, and without limiting the latent values to the range $[-z_{\max}, z_{\max}]$. We compare this version of the method to the original LAPO on four offline RL tasks that differ in the amount of high-performing task-relevant data contained in the offline dataset.

\subsection{Comparisons}
We compare LAPO to 7 offline RL approaches: 
(1) vanilla behavioral cloning (BC), 
(2) the BCQ method \cite{fujimoto2019off}, which approximates the dataset distribution with a generative model and manually selects the action with the maximum Q value among a set of generated actions, 
(3) the PLAS method \cite{zhou2020plas}, which learns a generative model via maximum likelihood and a latent-space policy to maximize the RL objective in the latent space,
(4) the AWAC method \cite{nair2020awac}, which performs advantage-weighted imitation learning with a forward-KL objective for the policy projection step,
(5) the IQL method \cite{kostrikov2021offline_iql}, which learns the value function using expectile regression without an explicit policy, and then extracts the policy using advantage weighted regression (AWR),
(6) the CQL method \cite{kumar2020conservative} which learns a conservative Q-function, 
and (7) AWAC\textsubscript{ {\tiny (GMM)}} which trains GMM policies using AWAC. 

In our experiments, we train each method three times using three different random seeds each for one million steps, 
and report the return averaged over 10 test episodes from the trained policies.
We used the implementation provided by \cite{seno2021d3rlpy} for the BCQ and AWAC, the original implementation for the PLAS, IQL and CQL, and our implementation for the BC method. We report the implementation details in the supplementary materials.

\begin{table*}
\vskip 0.15in
\caption{The normalized performance of all methods on tasks with heterogeneous dataset. 0 represents the performance of a random policy and 100 represents the performance of an expert policy. The scores are averaged over the final 10 evaluations and 3 seeds. LAPO achieves the best performance on 9 tasks and achieves competitive performance on the rest 3 tasks.}
\centering
\begin{small}
\begin{tabular}{lllllllll}
\toprule
Task ID                & BC & BCQ & PLAS & AWAC & AWAC\textsubscript{ {\tiny (GMM)}} & IQL & CQL & LAPO\textsubscript{{\tiny (Ours)}} \\
\midrule
Walker2d-mix-forward-v1   & -5.65  & 0.91   & 22.91         & 71.89           & 78.09     & 28.25           & \textbf{102.75}         & 74.17            \\
Walker2d-mix-backward-v1  & -84.56 & -1.37  & -27.58        & -7.95           & 71.33     & -46.25          & 66.64                   & \textbf{99.22}  \\
Walker2d-mix-jump-v1      & -72.92 & -41.43 & 0.56          & \textbf{51.08}  & 28.01     & -46.41          & 37.28                   & 43.20           \\ 
\midrule
Maze2d-umaze-v1           & 0.99   & 18.91  & 80.12         & 94.53           & 19.45     & 51.00           & 22.86                   & \textbf{118.86}  \\
Maze2d-medium-v1          & 3.34   & 12.79  & 5.19          & 31.40           & 46.53     & 33.26           & 12.25                   & \textbf{142.75}  \\
Maze2d-large-v1           & -1.14  & 27.17  & 45.80         & 43.85           & 9.04      & 64.30           & 7.00                    & \textbf{200.56}  \\ 
\midrule
Antmaze-umaze-diverse-v1  & 60.00  & 62.00  & 7.00          & 72.00           & 0.00      & 69.33           & 16.71                   & \textbf{91.33}   \\
Antmaze-medium-diverse-v1 & 0.00   & 11.33  & 8.67          & 0.33            & 0.00      & 73.00           & 1.00                    & \textbf{85.67}   \\
Antmaze-large-diverse-v1  & 0.00   & 0.67   & 1.33          & 0.00            & 0.00      & 48.00           & 11.89                   & \textbf{61.67}   \\ 
\midrule
Kitchen-complete-v0       & 4.50   & 9.08   & 38.08         & 3.83            & 1.08      & \textbf{66.67}  & 4.67                    & 53.17            \\
Kitchen-partial-v0        & 31.67  & 17.58  & 27.00         & 0.25            & 0.42      & 32.33           & 0.55                    & \textbf{53.67}   \\
Kitchen-mixed-v0          & 30.00  & 11.50  & 29.92         & 0.00            & 3.92      & 49.92           & 1.86                    & \textbf{62.42}   \\ 
\bottomrule
\end{tabular}
\end{small}
\label{table:multi-task}
\vskip -0.1in
\end{table*}
\raggedbottom

\begin{table*}[t]
\vskip 0.15in
\caption{The normalized performance of all methods on locomotion tasks with narrow and biased dataset. 0 represents the performance of a random policy and 100 represents the performance of an expert policy. The scores are averaged over the final 10 evaluations and 3 seeds. LAPO achieves the best performance on 4 tasks and achieves competitive performance on the rest of the tasks.}
\centering
\begin{small}
\begin{tabular}{llllllll}
\toprule
Task ID                & BC & BCQ & PLAS & AWAC & IQL & CQL & LAPO\textsubscript{{\tiny (Ours)}} \\

\midrule
Hopper-random-v2          & 2.23   & 7.80   & 6.68          & 8.01            & 7.89            & 8.33                    & \textbf{23.46}   \\
Walker2d-random-v2        & 1.11   & 4.87   & \textbf{9.17} & 0.42            & 5.41            & -0.23                   & 1.28             \\
Halfcheetah-random-v2     & 2.25   & 2.25   & 26.45         & 15.18           & 13.11           & 22.20                   & \textbf{30.55}   \\ 
\midrule
Hopper-medium-v2          & 49.23  & 56.44  & 50.96         & 69.55           & 65.75           & \textbf{71.59}          & 51.63            \\
Walker2d-medium-v2        & 47.11  & 73.72  & 76.47         & \textbf{84.02}  & 77.89           & 82.10                   & 80.75            \\
Halfcheetah-medium-v2     & 37.84  & 47.22  & 44.54         & 48.13           & 47.47           & \textbf{49.76}          & 45.97            \\ 
\midrule
Hopper-expert-v2          & 76.16  & 68.86  & 107.05        & \textbf{109.32} & \textbf{109.36} & 102.27                  & 106.76           \\
Walker2d-expert-v2        & 79.22  & 110.51 & 109.56        & 110.46          & 109.93          & 108.76                  & \textbf{112.27}  \\
Halfcheetah-expert-v2     & 85.63  & 93.15  & 93.79         & 14.01           & 94.98           & 87.40                   & \textbf{95.93}   \\
\bottomrule
\end{tabular}
\end{small}
\label{table:single-task}
\vskip -0.1in
\end{table*}
\raggedbottom

\subsection{Tasks and Datasets}
We compare LAPO to the introduced prior methods on learning from both heterogeneous and homogeneous offline datasets. 
We evaluate the methods on three simulated task domains, locomotion, navigation, and manipulation domains (Figure~\ref{fig:tasks}), with heterogeneous data distributions, and also a locomotion task with a narrow data distribution. 
We consider two sources of heterogeneous datasets: (1) data collected by policies accomplishing different tasks, and (2) data collected by different policies that accomplish one task but in different ways.  
For (1), we use the multi-task heterogeneous datasets introduced by \cite{yu2021conservative}, and for (2), we leverage heterogeneous datasets from the standard D4RL benchmark \cite{fu2020d4rl}. 
More details about our tasks and datasets are provided in the following: 

\begin{figure*}[ht]
\vskip 0.2in
\begin{center}
\centerline{\includegraphics[width=2\columnwidth]{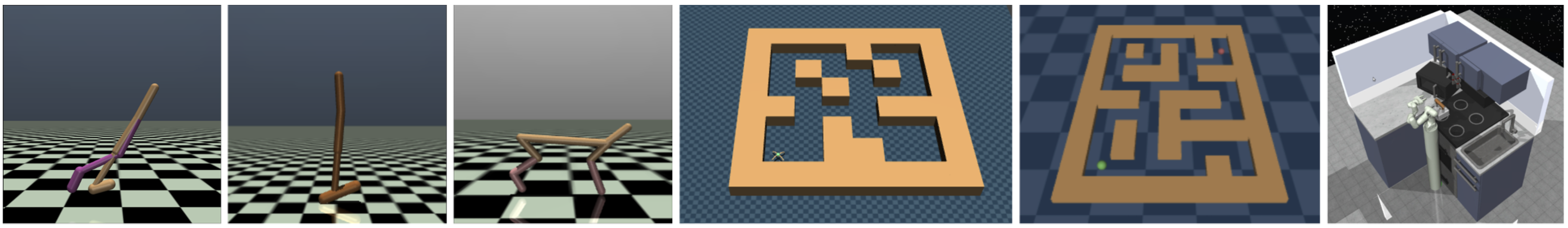}}
\caption{D4RL environments used in our evaluations: (from left to right) Walker 2D, Hopper, Halfcheetah,  Antmaze, Maze2d, and Kitchen environments. }
\label{fig:tasks}
\end{center}
\end{figure*}

\textbf{Locomotion:}  We adopt the task settings introduced in \cite{yu2021conservative} to define three locomotion tasks. The datasets are constructed using all of the training data in the replay buffer of three separate policy training sessions each trained for $0.5$ million steps using the SAC method \cite{haarnoja2018soft}. 
The tasks are to control a Walker2D agent to run forward, backward, and jump. 
Similar to the original work, we keep a single replay buffer with all of the transitions of all of the three tasks, and form three offline datasets by relabeling the rewards using the reward function provided for each task. We refer to the datasets as \textit{Walker2d-mix-forward}, \textit{Walker2d-mix-backward} and \textit{Walker2d-mix-jump} in the rest of this section.

\textbf{Navigation:} 
We use the two datasets \textit{Maze2d-sparse} and \textit{Antmaze-diverse} from the D4RL benchmark. The trajectories in the datasets are collected by training goal-reaching policies to navigate to random goals from random initial positions. 
Provided the pre-collected trajectories, the rewards are relabeled to generate offline data to navigate to different goal positions. Therefore, the task is to learn from data generated by policies that try to accomplish different tasks not aligned with the task at hand.
The target task has a sparse binary reward function which gives a reward of \emph{one} only when the agent is close to the goal position, and \emph{zero} otherwise. 

\textbf{Manipulation:} For the manipulation domain, we leverage the FrankaKitchen task from the D4RL benchmark. The task is to control a 9-DoF Franka robot to manipulate common household items such as microwave, kettle, and oven, in sequence to reach desired target configuration for several items. 
There are three datasets collected by human demonstrations: (1) \textit{Kitchen-complete} which consists of successful trajectories that perform tasks in order, (2) \textit{Kitchen-partial} which similar to (1) consists of some successful task-relevant trajectories, but also contains unrelated trajectories that are not necessarily related to reach any target configurations, and (3) \textit{Kitchen-mixed} which consists of partial trajectories that do not solve the entire task, and requires the highest level of generalization from the agent to accomplish the task. 
The kitchen environment has a sparse reward that is provided whenever an item is at its target configuration.

\textbf{Locomotion (narrow data distribution):}
For the offline RL task of learning from narrow data distributions,
we leverage three Gym-MuJoCo locomotion tasks from the D4RL benchmark with narrow and biased data distribution: \textit{Hopper}, \textit{Walker2d} and \textit{Halfcheetah}. Each task contains three datasets collected using a randomly initialized policy ("-random"), a semi-trained policy ("-medium"), and a fully trained policy ("-expert"), respectively. The behavioral policies are trained online using the SAC method.

\begin{figure}[h]
\vspace{-0.8cm}
\begin{center}
\centerline{\includegraphics[width=0.8\columnwidth]{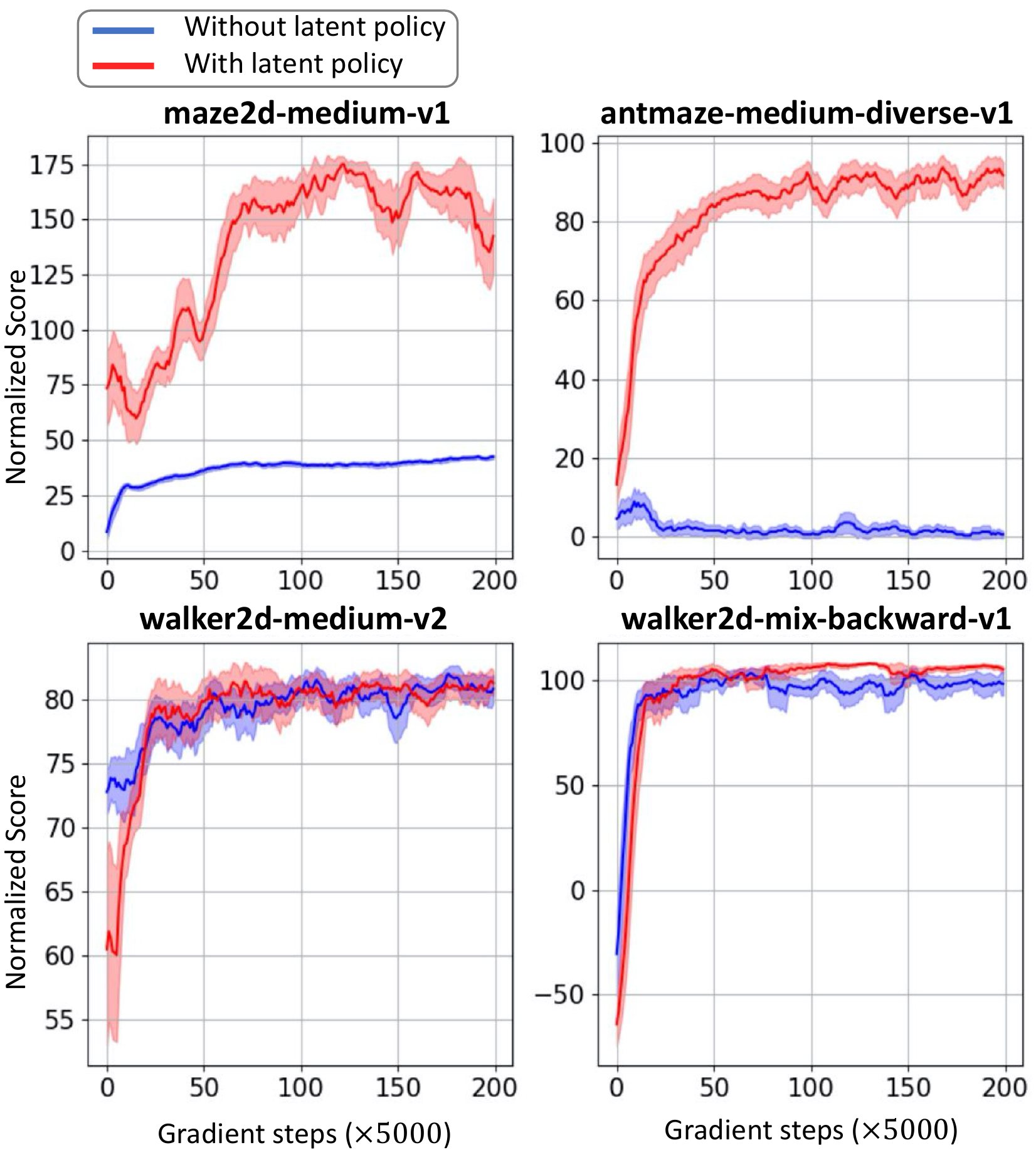}}
\caption{Comparison of LAPO with and without the latent policy on Maze2D-medium, Antmaze-medium-diverse, Walker2D-medium, and Walker2D-mix-backward task. 
The shaded area represents ±95\%-confidence intervals, computed using 10 evaluations and 3 random seeds. 
For tasks with less optimal task-relevant data like Maze2D-medium, Antmaze-medium-diverse, the learned latent policy results in high-performance gains.}
\label{fig:ablat_latentpi}
\end{center}
\vskip -0.2in
\end{figure}

\subsection{Experimental Results}
\textbf{Heterogeneous datasets:} Table~\ref{table:multi-task} reports the results of our experiments on heterogeneous datasets\footnote{\label{fn: std}Results with 95\%-confidence interval are reported in the Appendix.}. Among the 12 tasks introduced in the previous section, LAPO (our method) achieves the best performance on nine tasks and the second-best performance on one task.
Besides, on average, LAPO improves by 49\%  over the next best method on the heterogeneous datasets.

The heterogeneous locomotion tasks can be accomplished more easily by the prior methods since still $33\%$ of the data comes from the target task. 
However, from the prior works, AWAC and CQL are the only methods that can accomplish the tasks. 
LAPO performs well on all tasks and on average yields the best performance on one out of the three tasks. 
This shows that LAPO is capable of learning from heterogeneity introduced by multi-task datasets. 

The navigation tasks are in general more challenging especially for those with medium and large map sizes. The main challenge is to learn policies for long-horizon planning from datasets that do not contain optimal trajectories, and there are only very few states with rewards.  
LAPO significantly outperforms all prior works for all of the navigation tasks. 
Similarly, the manipulation tasks are also challenging, as they require the assembly of sub-trajectories related to completing a given task consisting of multiple sub-tasks. In addition, the agent has access to fewer training samples while having to learn to interact with complicated dynamics.  
LAPO outperforms previous approaches by a large margin on Kitchen-partial and Kitchen-mixed for which only a few optimal trajectories are provided by the dataset. It also performs competitively on the Kitchen-complete task and yields the second-best performance. 

Surprisingly, we observe that GMM policy training does not perform well on heterogeneous offline RL settings. The AWAC method with GMM policies yields similar or even worse performance compared to the original AWAC. 
These results suggest that latent variable models are better candidates to effectively model the high-reward regions of data especially when given data with multiple modes. 

\textbf{Narrow datasets:}
Table~\ref{table:single-task} reports the results of the locomotion tasks with narrow and biased datasets. Our method yields the best performance for four out of nine tasks. 
It also achieves competitive performance on the rest of the tasks, showing that LAPO is applicable to general offline RL settings and is not limited to learning from heterogeneous datasets. However, the most significant gain is when it is applied to offline settings with heterogeneous datasets.

\subsection{Ablations}
\textbf{Latent Policy Training}:
To study the importance of the latent policy training in achieving good performances, 
we conduct an ablation study on $4$ different tasks by eliminating the latent policy training from LAPO.
The main motivation is that, 
as described in Section ~\ref{sec:method}, the action policy $\bpi$ is trained to approximate the optimal policy $\pi^*$, hence training the latent policy on top of that may seem redundant. In this case, another option for sampling latent variables is to sample from the prior distribution, which changes Equation \ref{eq:value_function} to: 
\begin{equation*}
    \label{eq:value_function_vaeaction}
    V(s) = \E_{ \substack{ a \sim \pi_\theta(. | s, z) \\ z \sim p(z) }} [Q_{\phi'}(s, a)],
\end{equation*}
where, the latent $z$ is now sampled from a fixed prior.

As shown in Fig.~\ref{fig:ablat_latentpi}, training a latent policy does not significantly contribute to higher performance for tasks such as the Walker2D-medium and Walker2d-mix-backward which contain sufficient task-relevant data.
However, latent policy training results in large performance gains when learning from datasets which contain fewer optimal task-related data such as Maze2D and Antmaze. 
LAPO overcomes this challenge by explicitly optimizing the RL objective through learning the latent policy.

\begin{figure}[h]
\vskip 0.2in
\begin{center}
\centerline{\includegraphics[width=0.8\columnwidth] {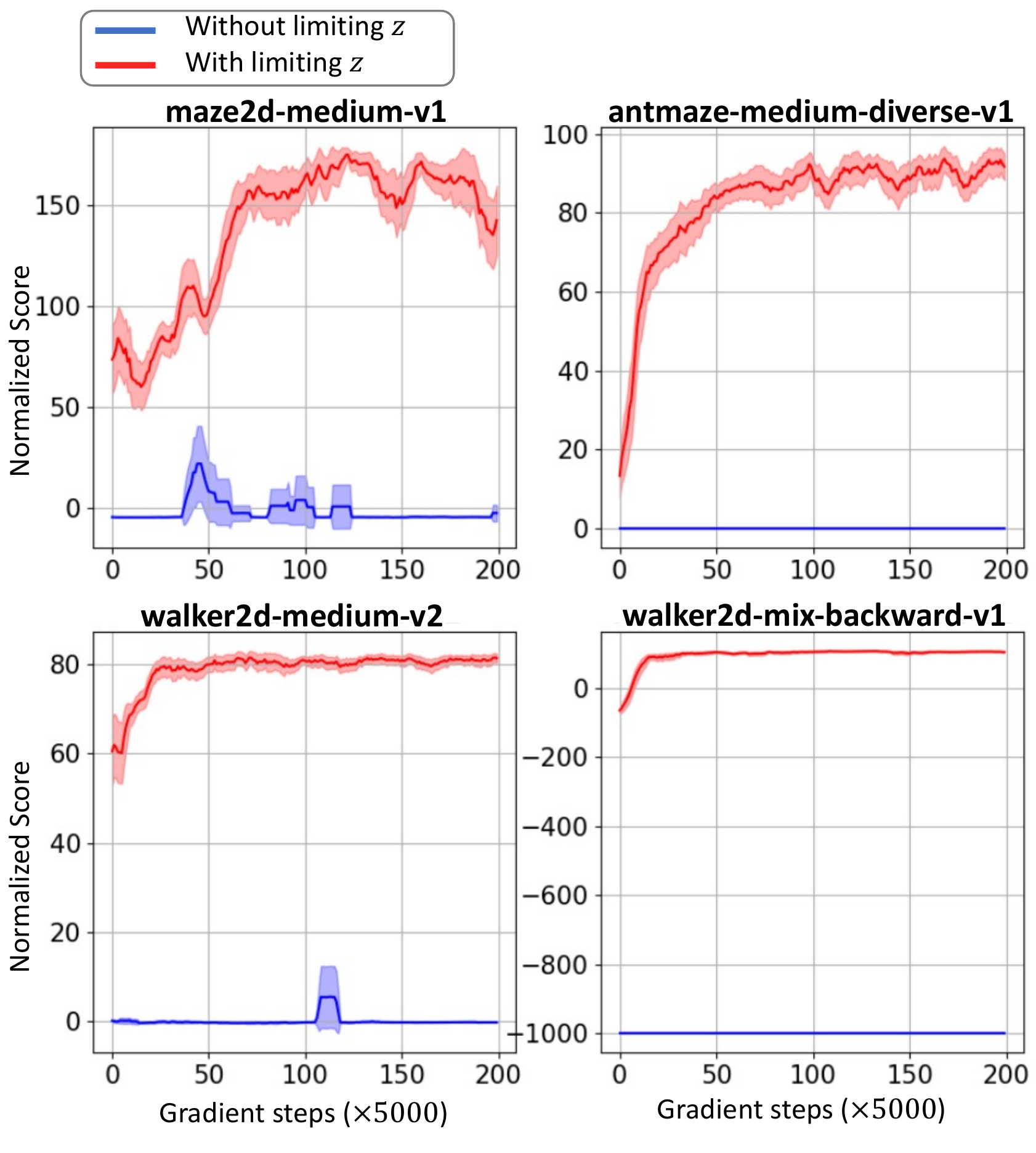}}
\caption{Comparison of LAPO with and without limiting the latent values.  
The policy performance is significantly worse when the latent values are not limited to the range $[-z_{\max}, z_{\max}]$. }
\label{fig:z_limit}
\end{center}
\vskip -0.2in
\end{figure}

\textbf{Limiting the latent values}:
We also study the importance of limiting the latent variable $z$ to be within the range $[-z_{\max}, z_{\max} ]$. This is important to understand the limits of using the action policy as a generative model of in-distribution actions. Figure~\ref{fig:z_limit} illustrates the result of LAPO policy training using unbounded latent actions on four tasks. The policy training performance is significantly worse when we do not limit the latent values. This suggests that the the action policy can generate in-distribution actions only when in-distribution latent values (close to samples drawn from the prior distribution $p(z)$) are given as the input. 

\section{Conclusion}
In this paper, we study an offline RL setup for learning from heterogeneous datasets where trajectories are collected using policies with different purposes, leading to a multi-modal data distribution. Through empirical analysis, we find that in such cases, policies constrained with forward-KL or MMD may contain out-of-distribution actions and lead to suboptimal performance for continuous control tasks, especially for heterogeneous datasets.
To address this challenge, we present the latent-variable advantage-weighted policy optimization (LAPO) algorithm, which learns a latent variable policy that generates high-advantage actions when sampling from a prior distribution over the latent space. 
In addition, we train a latent policy that obtains state-conditioned latent values which result in higher reward outcomes compared to sampling from the prior distribution.
We compare our method to 6 prior methods on a variety of simulated locomotion, navigation, and manipulation tasks provided heterogeneous offline datasets, and also on standard offline RL benchmarks with narrow and biased datasets. 
We find that our proposed method consistently outperforms prior methods by a large margin on tasks with heterogeneous datasets, while being competitive on other offline RL tasks with narrow data distributions.
For our future work, we will extend LAPO to multi-task offline RL settings in which an agent learns multiple RL tasks provided a heterogeneous dataset of diverse behaviors.

\bibliographystyle{bib_sobraep}
\bibliography{referencias_sobraep}

\end{document}